\documentclass[journal]{IEEEtran}

\ifCLASSINFOpdf
\usepackage{graphicx}
\usepackage{epstopdf}
 \else

\fi
\usepackage{amsmath}
\usepackage[ruled,vlined]{algorithm2e}
\usepackage[noend]{algpseudocode}
\usepackage{stfloats}
\usepackage{caption}
\usepackage{cite}
\usepackage[numbers,sort&compress]{natbib}
\usepackage[numbers]{natbib}
\usepackage{array}
\usepackage{subfigure}
\usepackage{amsmath,amsthm}
\usepackage{amsmath,bm}
\usepackage{amstext,amssymb}

\theoremstyle{definition}

\begin{document}
\title{Mask-FPAN: Semi-Supervised Face Parsing in the Wild With De-Occlusion and UV GAN}





\author{
\IEEEauthorblockN{
Lei Li\IEEEauthorrefmark{1}, 
Tianfang Zhang\IEEEauthorrefmark{2}, 
Zhongfeng Kang\IEEEauthorrefmark{1}, and
Xikun Jiang\IEEEauthorrefmark{1} \\
\IEEEauthorblockA{\IEEEauthorrefmark{1}Department of Computer Science, University of Copenhagen, Denmark\\ lilei@di.ku.dk \\}
\IEEEauthorblockA{\IEEEauthorrefmark{2}School of Information and Communication Engineering, University of Electronic Science and Technology of China\\}
} 
}

\maketitle
\begin{abstract}
The field of fine-grained semantic segmentation for a person's face and head, which includes identifying facial parts and head components, has made significant progress in recent years. However, this task remains challenging due to the difficulty of considering ambiguous occlusions and large pose variations. To address these difficulties, we propose a new framework called Mask-FPAN. Our framework includes a de-occlusion module that learns to parse occluded faces in a semi-supervised manner, taking into account face landmark localization, face occlusionstimations, and detected head poses. Additionally, we improve the robustness of 2D face parsing by combining a 3D morphable face model with the UV GAN. We also introduce two new datasets, named FaceOccMask-HQ and CelebAMaskOcc-HQ, to aid in face parsing work. Our proposed Mask-FPAN framework successfully addresses the challenge of face parsing in the wild and achieves significant performance improvements, with an MIOU increase from 0.7353 to 0.9013 compared to the current state-of-the-art on challenging face datasets.
\end{abstract}

\IEEEpeerreviewmaketitle

\section{Introduction}
Over the past few years, there have been successful applications of face detection, human facial landmark localization, and face parsing schemes in various fields \cite{jackson2016cnn, luo2012hierarchical}. However, labeling processes for these tasks can be complex and time-consuming. To address these issues, Wayne \cite{wu2018look}, Liu \cite{liu2015multi}, and Lee \cite{lee2020maskgan} have developed solutions to resolve ambiguities in labeling facial landmarks and face parsing, thereby improving the labeling process.

Despite recent advancements, there still exist several challenges in the field of face parsing. Firstly, labeling fine-grained facial parts can be a disturbance, as obtaining unambiguous datasets through effective and easy labeling is difficult. Additionally, efficient face segmentation requires more feature learning through complex methods. Furthermore, it is worth mentioning that face parsing performance is often inadequate in real-world scenarios, particularly when faced with large pose variations and diverse occlusion objects.

Our method takes inspiration from human receptive attention in identifying faces, where humans not only focus on the principal facial features but also objects obscuring the face. However, previous work has not focused specifically on occluded facial features for face parsing, whether through object occlusion or self-occlusion at a large angle. In line with Occam's razor principle, we have developed a lightweight network to address the challenges of face parsing in real-world scenarios with actual occlusions, large angles, and other complex scenarios. This approach contrasts with current methods that work with unoccluded datasets. Our method places a higher value on complex scenarios, and we utilize a semi-supervised augmentation process.

\begin{figure*}[ht]
\centering
\includegraphics[width=0.8\textwidth]{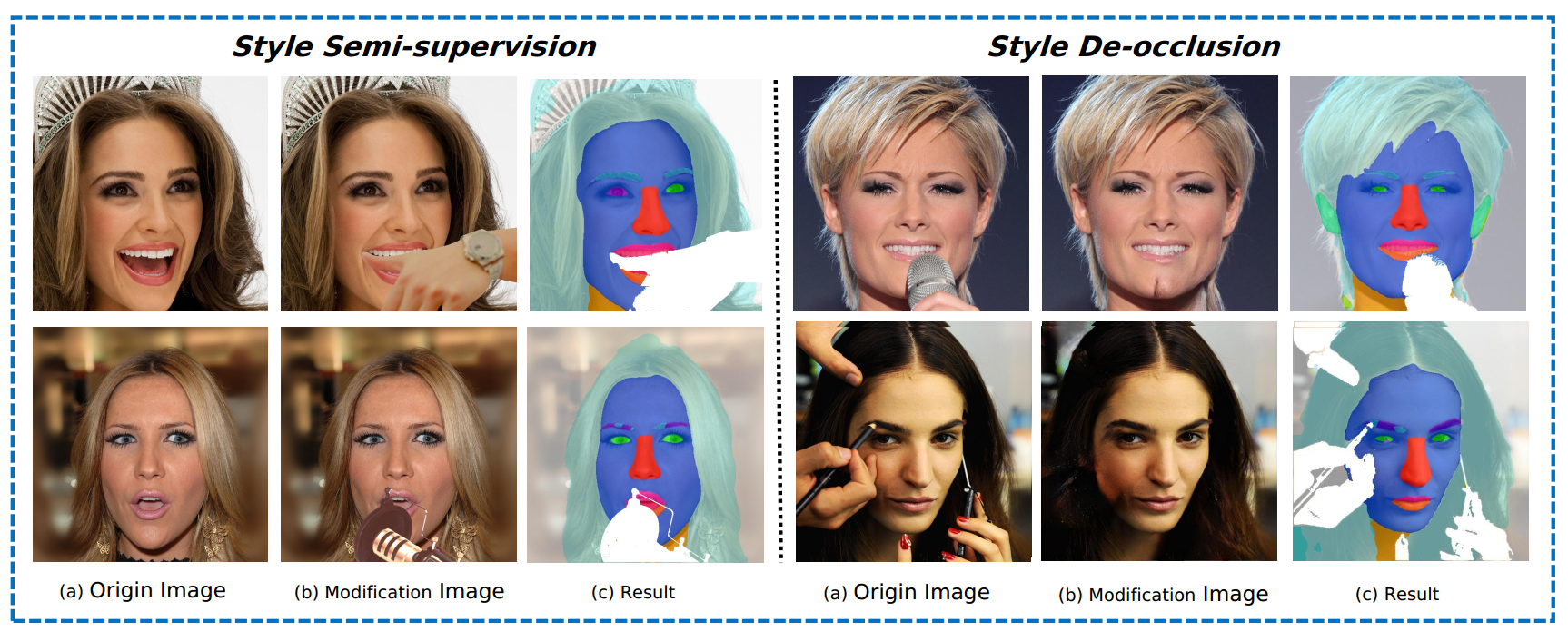}
\caption{We propose a method for tackling the challenge of face parsing in the wild, which accounts for occlusions and large pose variations. Our method achieves superior performance in semantic segmentation of facial components and is capable of handling occluded faces and large pose variations. A semi-supervised sub-model modifies the original image, while a style de-occlusion sub-model removes such modifications. Additionally, a UV GAN module, combined with a UV mask, enhances the robustness of face parsing from large angles. The proposed methods collectively contribute to improved face parsing performance. (Best viewed in color)}
\label{fig:teaser}
\end{figure*}

We propose a complete pipeline that includes (1) a semi-supervised sub-model combining face-dense landmarks prediction, (2) a face occlusion prediction sub-model, (3) a de-occlusion sub-model with face, and (4) UV map completion of face masks in the training process to ensure facial part segmentation. It is important to note that while the segmentation module is a part of our framework, it is not the main focus of our work. Thus, we have chosen the most efficient and straightforward network for running on embedded machines.

There are only a few publicly available face parsing datasets, including HELEN~\cite{smith2013exemplar}, LFW~\cite{kae2013augmenting}, and CelebAMask-HQ~\cite{lee2020maskgan}. However, HELEN and LFW datasets have limited class distinctions and do not include hair, occlusion, or other extracorporeal elements. In contrast, CelebAMask-HQ is a large-scale high-resolution face dataset with fine-grained mask annotations, but it lacks occlusion and varying poses. To address these limitations, we propose two new datasets, FaceOccMask-HQ and CelebAMaskOcc-HQ, which include more face parts, occlusion elements, and arbitrary poses. To facilitate face parsing in the wild, we manually collected face parsing training data and demonstrate that our network outperforms state-of-the-art results on both FaceOccMask-HQ and CelebAMaskOcc-HQ datasets.

The work originated from a real-world problem: how to achieve real-time and effective face parsing on a mobile phone in the wild? This involves ensuring a lightweight segmentation network while enabling end-to-end learning of the entire training framework. Rather than improving face parsing on a specific dataset using a complex segmentation model, our focus is on the overall performance in real-world scenarios. It should be emphasized that the novelty of our method lies in the following aspects.
\begin{enumerate}
\item We propose a novel method to handle the problem of face parsing in the wild by leveraging various techniques. Firstly, we propose and verify the effectiveness of using 3D face features to improve the face parsing performance. To the best of our knowledge, it is the first attempt to integrate facial landmark localization, face occlusion estimation, face pose estimation, and 3D morphable model with UV GAN for face parsing. The proposed method effectively addresses the sparse relationships between different facial parts and leads to a significant performance improvement, as demonstrated in Figure~\ref{fig:teaser}.
 
\item Both the generated and annotated datasets, FaceOccMask-HQ and CelebAMaskOcc-HQ, are of higher quality and are better suited for face parsing research. The datasets have been validated to contain more complex scenarios, including various occlusions and arbitrary large-angle poses encountered in the wild.

\item Two novel semi-supervised methods are employed for the first time in the face parsing training process.
\subitem We have developed a novel de-occlusion module called Occ-Autoencoders, which significantly improves performance when occluding components such as hands, masks, glasses, necklaces, and microphones are present in front of faces. During the training process, we use the de-occlusion module to automatically generate masks for occluded parts when corresponding masks are not available. This enables semi-supervised learning for the entire face parsing process, allowing the system to handle such scenarios even without labels.
 
\subitem We propose a novel approach where we combine a UV GAN module with a UV mask to enhance the robustness of face parsing from large angles. This allows us to automatically generate mask labels without the need for extensive manual labeling efforts, either online or offline. By leveraging this approach, we can significantly improve the overall performance of face parsing, while also contributing to the 3D face part segmentation when faces are reconstructed in 3D.
\end{enumerate}

\begin{figure*}[htb]
\centering
\includegraphics[scale=0.15]{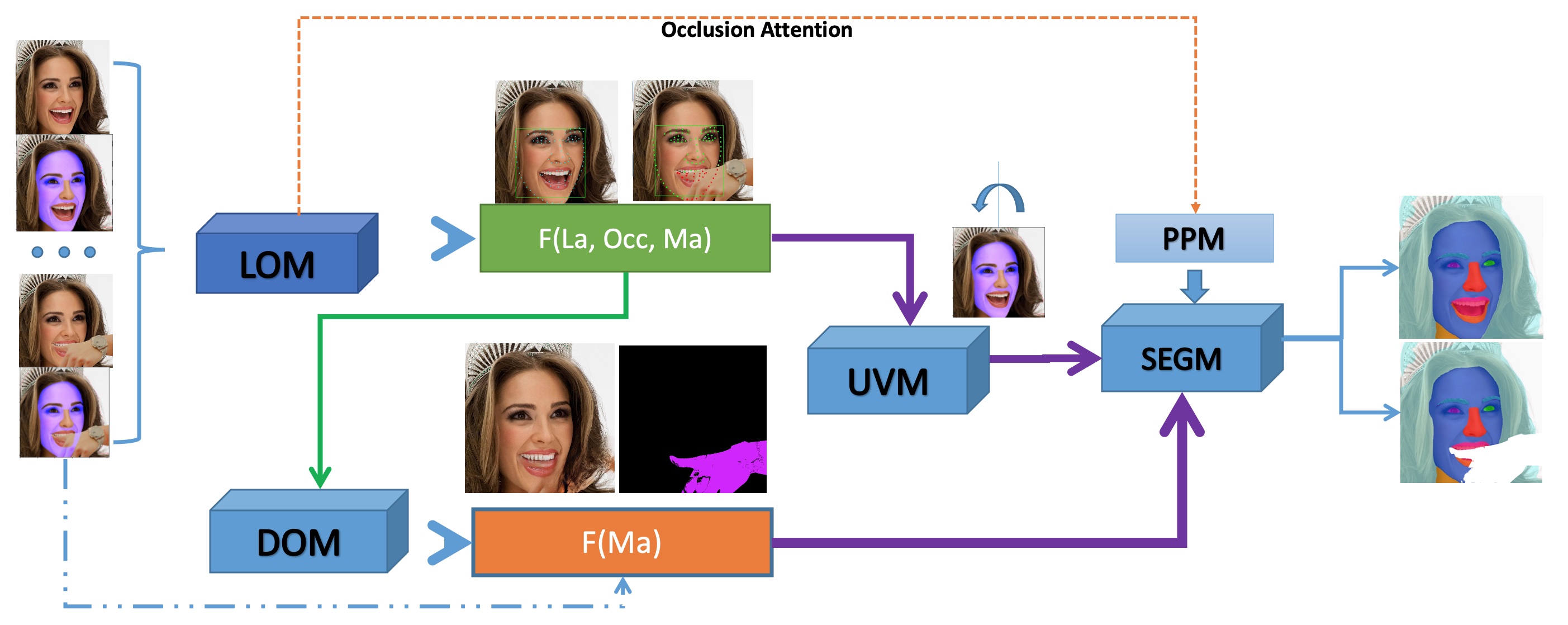}
\caption{An overview of our framework for face parsing is shown. It can be divided into pre-trained stage and training stage. In the pre-trained stage, \textsl{LOM}, \textsl{DOM}, \textsl{UVM} and \textsl{PPM} are separately trained, then the trained data will first be loaded in the \textsl{LOM} on the foundation of results from the \textsl{LOM}, which are determined whether the data is going into the \textsl{DOM} or not. If there is no occlusion detected on the face, it is directly sent to the \textsl{UVM}, which will then use it to create random poses of the face and correspondent segmentation maps automatically; otherwise, it goes to the \textsl{DOM} to remove the occlusion, return the occlusion mask, and finally combine the original facial part mask to the next \textsl{SEGM}. This is the semi-supervised sub-model of our framework. Furthermore, in the training stage, we use the mask labels and images from UVM as inputs to the \textsl{SEGM} and the former results from \textsl{LOM} and \textsl{DOM} with attention from \textsl{PPM} will assist the whole training process. In the test process, we just operate \textsl{SEGM} for face parsing prediction.}
\label{framwork}
\end{figure*}

\section{Related Work}

\textbf{Facial landmark localization} has become a crucial topic in computer vision. It has various embedded applications, such as face beauty and face unlocking. Researchers have been focusing on the performance and robustness of facial landmark localization, like \cite{wu2018look,Watchareeruetai2022,guo2019pfld,dong2018style,Xia2022Splandmark,feng2018wing,wang2019adaptive,wang2019analysis}. PFLD \cite{guo2019pfld} performs landmark detection for different poses, expressions, and occlusions on mobile phones, and we use it for our landmark location prediction. LAB \cite{wu2018look} handles challenging cases by using the facial boundary. SAN \cite{dong2018style} adds data augmentation by a generative adversarial module, and SBR \cite{dong2018supervision} utilizes the fact that objects move smoothly in a video sequence to improve an existing facial landmark detector. Additionally, loss function design \cite{feng2018wing,wang2019adaptive} has been proposed to achieve significant improvements on various evaluation metrics.

We evaluated several previous works \cite{wu2018look,jackson2016cnn,guo2019pfld,dong2018style,burgos2013robust,feng2018wing,wang2019adaptive} and adopted the PFLD \cite{guo2019pfld} approach, which is a lightweight network capable of predicting face landmarks and occluded face parts. We then incorporated landmark occlusion prediction modules (see Figure~\ref{occ_results}) to further enhance the performance of face parsing.


\textbf{Semantic Segmentation}, an elementary computer vision task for 2D generic images, has achieved impressive results with various approaches \cite{chen2017deeplab,Li2022building,yu2018bisenet,wang2022rtformer,10024907,kirillov2023segment}. One of the most successful approaches for semantic instance segmentation is Mask-R-CNN \cite{he2017mask}, which is based on the detect-then-segment strategy. Another approach is to label each pixel of an image and subsequently add auxiliary information that a clustering algorithm can use to group pixels into object instances \cite{long2015fully,chen2017deeplab,yu2018bisenet}. As opposed to using huge networks for semantic segmentation, real-time performance on mobile embedded applications is of more concern, and this has led to the development of diverse approaches for accelerating the operation running time (e.g., \cite{badrinarayanan2017segnet,li2017not,paszke2016enet,zhao2018icnet}).
Although a big general model like \cite{kirillov2023segment} can be used to segment multiple objects, we are concerned with real-time performance when using embedded equipment.

Recent research has investigated various approaches to \textbf{Face Parsing}, which can be broadly categorized into global \cite{wei2017learning, saito2016real, te2021agrnet} and local methods \cite{lin2019face, luo2012hierarchical, zhou2015interlinked}. With the advancements in deep learning techniques, several convolution neural network structures and loss functions have been proposed to encode the underlying layouts of the whole rigid face. Local methods typically involve training separate models, such as RoIAlign methods \cite{lin2019face}. Given our focus on occlusions, we have implemented a local method with an attention mechanism for occlusion components, along with a global method for other modules, to meet real-time performance requirements.

The \textbf{Face De-Occlusion} module is essential for obtaining satisfactory results in face alignment and parsing, especially in occlusion scenes. Whole de-occlusion methods \cite{burgos2013robust,zhao2017robust,zhang2016occlusion,Li2023Multi} often involve image inpainting and feature fusion\cite{bertalmio2000image,Wu2019FASE,liu2018image,ZhangFact2020,Zhang2022LR-CSNet,Jaime2022Fusion}. Among them, the method proposed in \cite{zhao2017robust} has shown superior performance compared to other de-occlusion methods, especially when dealing with partially occluded faces.

We incorporate landmark localization into our approach to create occlusion masks for facial parts based on the landmark positions, as shown in Figure~\ref{CeleAMaskOcc}. This helps improve the robustness of our de-occlusion module, which does not require exact locations of corrupted regions, as illustrated in Figure~\ref{de-occlusion_moudle}. To the best of our knowledge, our framework is the first to use de-occlusion methods for semi-supervised face parsing, automatically identifying occluded regions and creating occlusion masks that improve segmentation performance. The \textbf{UV GAN} module \cite{deng2018uv} is a global and local adversarial network that learns identity-preserving UV completion. We leverage this module by training it on faces in different directions corresponding to occluded UV textures and the whole UV texture map. When we attach the completed UV map to the fitted 3D mesh, we can use the UV GAN module to randomly rotate the frontal face and correspondent facial part mask, obtaining faces and face parsing masks for arbitrary poses. This increases pose variation during training and improves face parsing performance for large postures.

\section{Method}
Our aim is to achieve more accurate and robust face parsing results in unconstrained environments. The proposed framework, illustrated in Figure~\ref{framwork}, comprises multiple modules, namely the \textsl{Landmark Occlusion Module} (\textbf{LOM}) discussed in Section~\ref{LOM}, the \textsl{De-occlusion Module} (\textbf{DOM}) explained in Section~\ref{DOM}, the \textsl{UV GAN Module} (\textbf{UVM}) described in Section~\ref{SectionUVM}, the \textsl{Pose Prediction module} (\textbf{PPM}) detailed in Section~\ref{PPM}, and the \textsl{Segmentation Module} (\textbf{SEGM}) elaborated in Section~\ref{SEGM}. Each of these modules will be further explained in the subsequent sections.

Assuming we have a combined training set for semi-supervised training, denoted by ${{(x_i, L_i)}^N_{i=1}}$, which contains N face images $x_i$ and their corresponding feature mask labels $L_i$. The target image is denoted as $I^t \in \mathbb{R}^{H\times W\times 3}$. The dataset contains three types of labels: (1) masks for occluded faces, (2) faces with occluded components but without mask labels, and (3) naked faces.

We propose using the \textsl{LOM} to identify occluded regions and determine the locations of facial landmarks. If all occluded regions are marked in the occlusion masks or there are no occlusions, semantic segmentation can be performed directly. However, if there are occlusions, the \textsl{DOM} will be utilized to remove the occlusions and obtain the occlusion mask label. It is worth mentioning that the original datasets, such as FaceOccMask-HQ and CelebAMaskOcc-HQ, mostly consist of frontal face images, hence the \textsl{UVM} is employed to generate high-fidelity face datasets with varying poses. These modules can be utilized either during the training process or in preprocessing the datasets, thus allowing for semi-supervised face parsing of the entire scene.

Our proposal involves a face parsing framework based on a segmentation network. This framework takes into account occluded regions as well as pose variations, and involves fine-grained semantic segmentation.

\begin{figure}[!ht]
\centering
\includegraphics[scale=0.13]{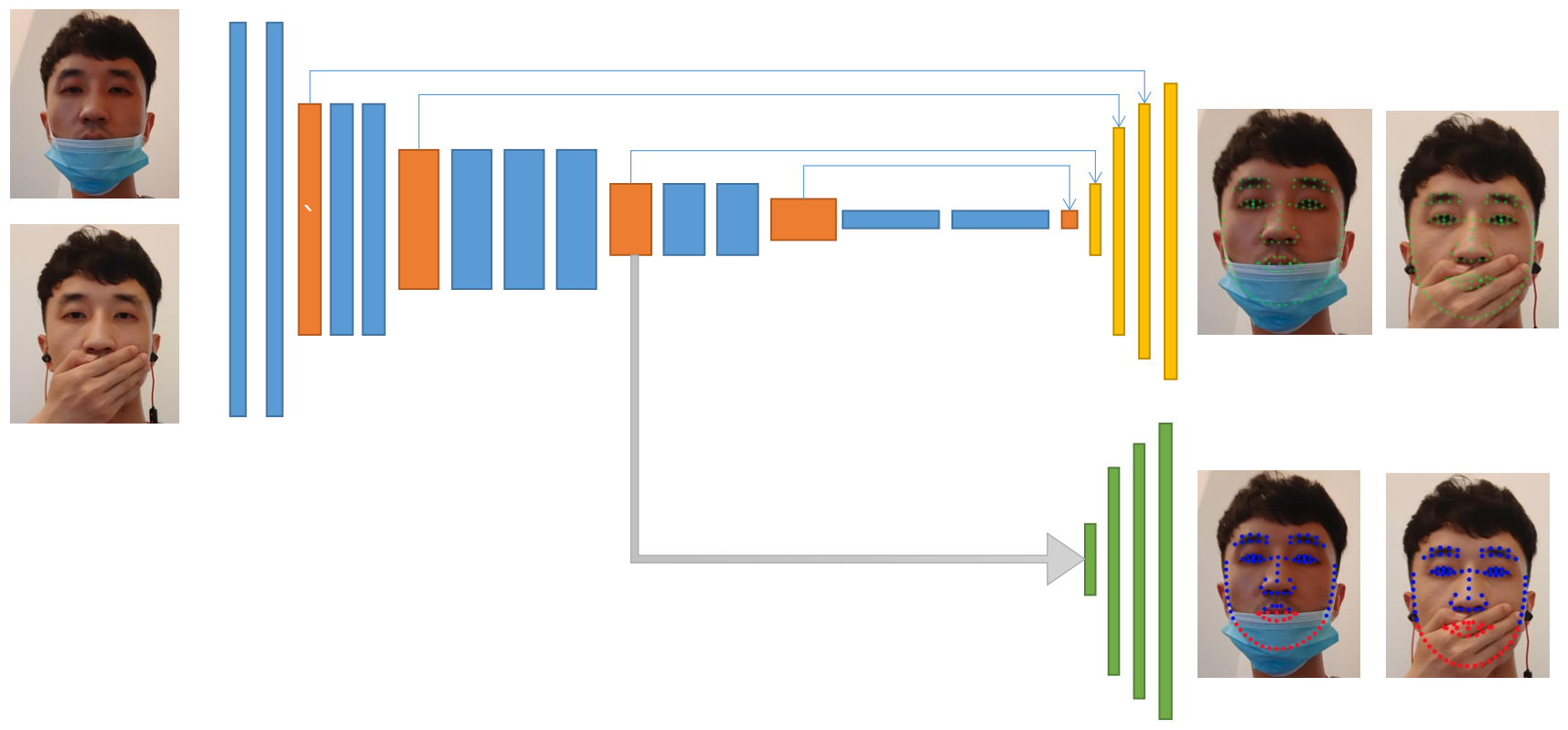}
\caption{Landmark occlusion module. This is used for 3D fitting and occlusion prediction.}
\label{landmark_occ_module}
\end{figure}

\begin{figure}[ht]
\centering
\includegraphics[scale=0.18]{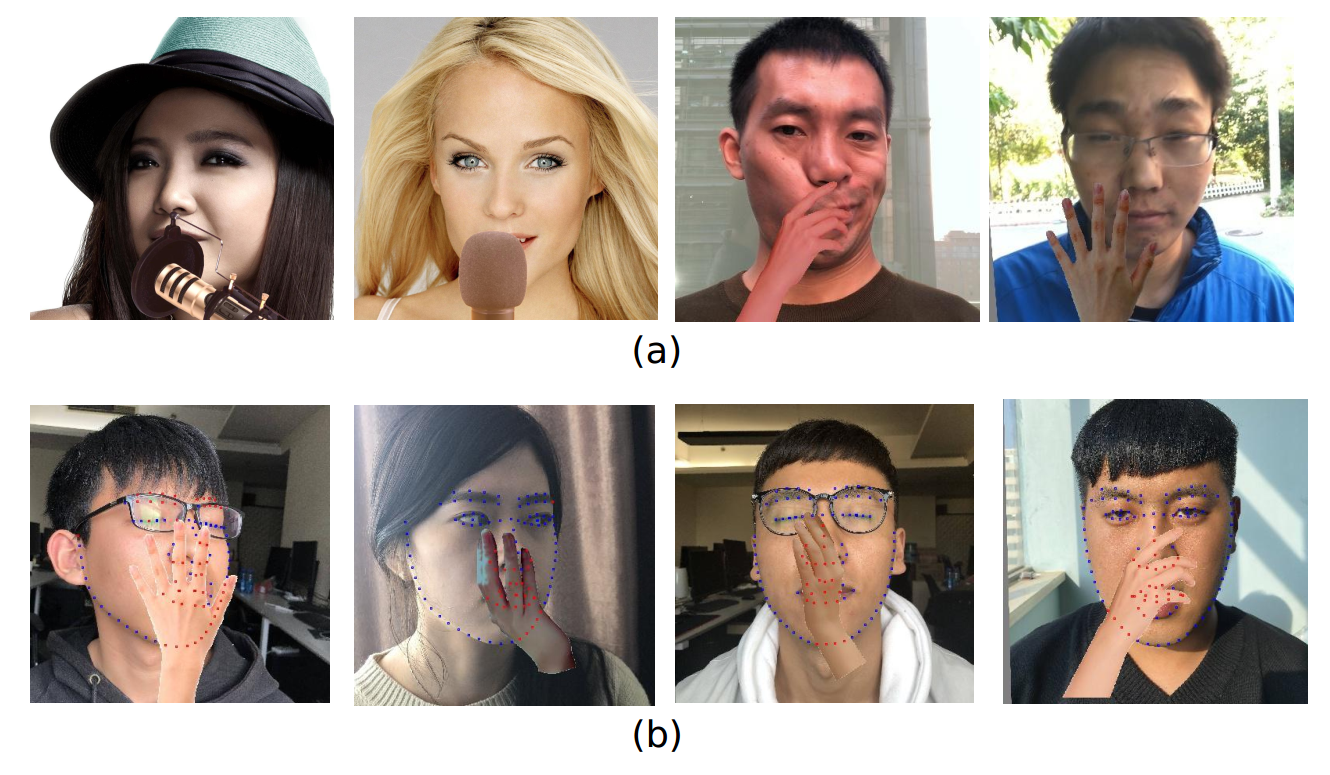}
\caption{Samples of our landmark occlusion module training dataset (a) and {b}. Our data covers common objects, which gives efficient effects when we estimate the occlusion details.}
\label{occ_train_dataset}
\end{figure}

\begin{figure}[!ht]
\centering
\includegraphics[scale=0.28]{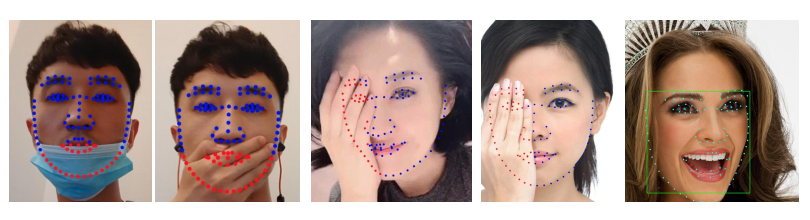}
\caption{Some examples show multiple occlusions in the face and the LOM can be used to judge which regions are occluded.}
\label{occ_results}
\end{figure}

Our proposed framework utilizes the input from the \textsl{LOM}, which gives more weight to related parts during the training process when those parts are occluded. This weight adjustment can be found in the \textsl{SEGM}. It should be noted that while the \textsl{LOM}, \textsl{DOM}, \textsl{UVM}, \textsl{PPM}, and \textsl{SEGM} are all used during the training process, we only use the \textsl{SEGM} to predict face parsing in order to meet the real-time requirements of embedded devices.

\begin{figure*}[ht]
\centering
\includegraphics[scale=0.27]{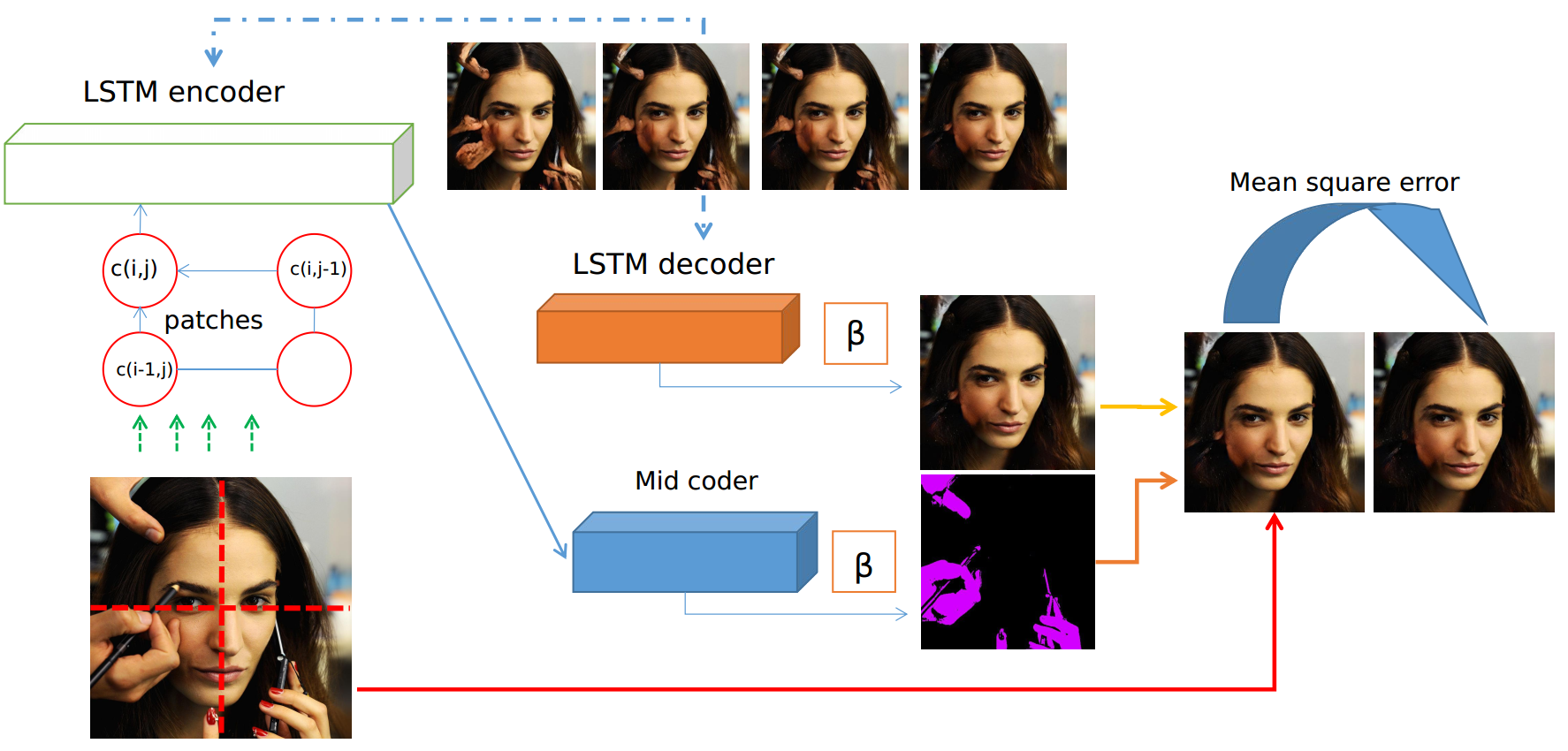}
\caption{The details of the de-occlusion module. It consists of a multi-scale spatial LSTM encoder and a dual-channel LSTM decoder for concurrent face reconstruction and occlusion detection.}
\label{de-occlusion_moudle}
\end{figure*}

In our framework, we use two functions, $F(\text{landmark}, \text{Occlusion}, \text{Mask})$ and $F(\text{Mask})$, to train in an end-to-end manner by combining all the modules. The $F(\text{landmark}, \text{Occlusion}, \text{Mask})$ function works on the \textsl{LOM} to determine occluded parts with landmark point occlusion, as seen in the occlusion landmark judgment (refer to Figure~\ref{occ_results}). To accurately segment occluded parts, we design a loss function that calculates the distance of other parts from the occlusion landmark using a proper distance threshold. When some face regions are occluded, they are often segmented with ambiguous boundaries, making it difficult to accurately divide the occluded parts. Additionally, the $F(\text{Mask})$ function is used to determine the existence of mask labels. If there are no mask labels with occlusion, the image set will be processed through the \textsl{DOM}, whose mechanisms are explained in Section~\ref{DOM} and illustrated in Figure~\ref{de-occlusion_moudle}. We will now discuss these modules in further detail.


\subsection{Landmark Occlusion Module} \label{LOM}
\begin{figure}[ht]
\centering
\resizebox{0.75\columnwidth}{!}{\includegraphics{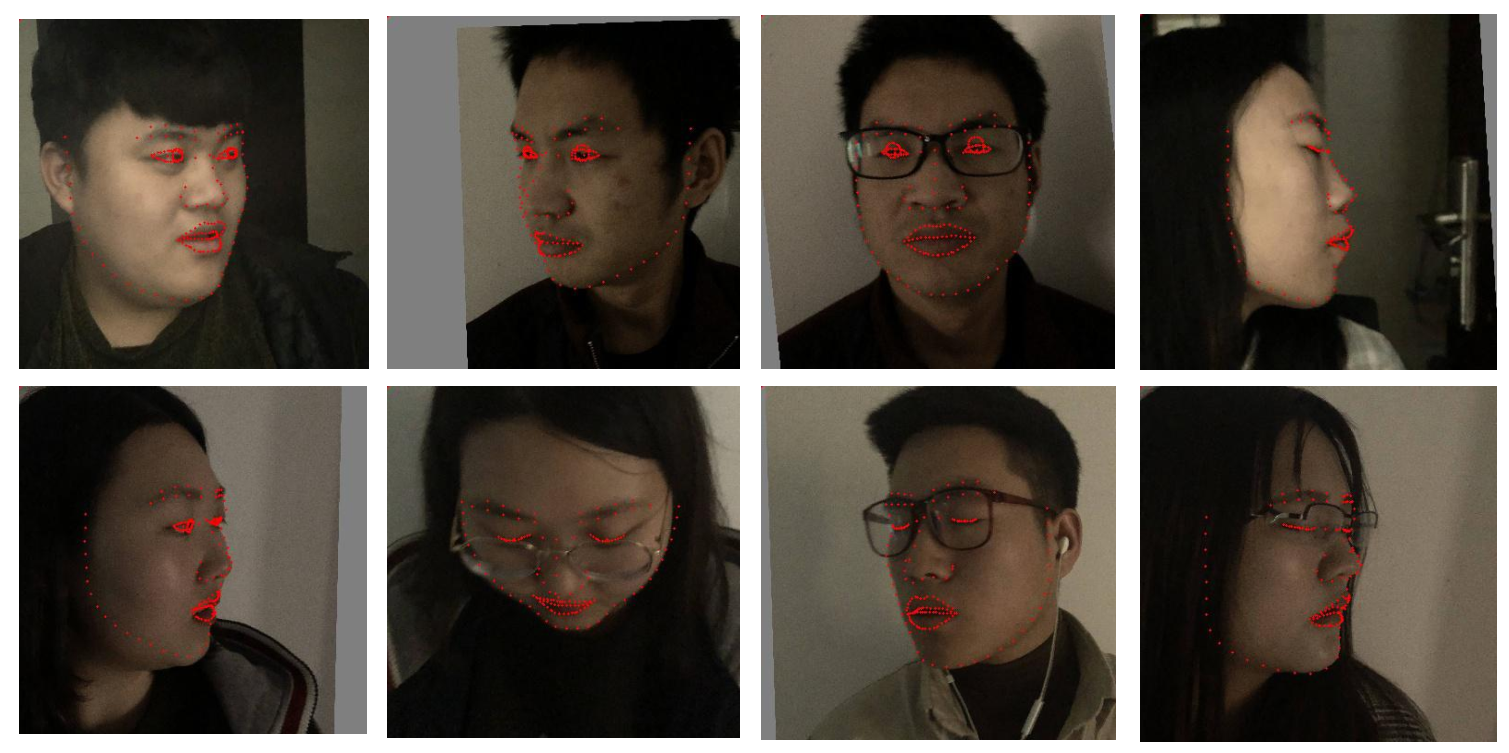}}
\caption{We use the dense points about 264 landmark points to illustrate our face parts. The more accurate the points, the more efficient is the face parsing we can perform.}
\label{264point}
\end{figure}

Previous face parsing works have often used landmark labels to obtain part labels for segmentation. In contrast, we define dense landmarks with 264 points (as shown in Figure~\ref{264point}) and use the main points for occlusion prediction training (as seen in Figure~\ref{occ_train_dataset}) on faces in the wild. These dense landmarks allow us to crop facial parts for subsequent face parsing. Additionally, we combine facial landmark occlusion status (as illustrated in Figure~\ref{landmark_occ_module}) in a novel way to facilitate the intersection between occlusions and areas close to occluded parts. More details on this approach can be found in Section~\ref{SEGM}.

To elaborate, we obtain landmark localization predictions in the first stage. Subsequently, we fix this branch and train the occlusion prediction branch using our occlusion train dataset. We leverage our dense points to acquire semantic parts of the face and boundary points to construct our training dataset. In this manner, we use facial parts landmarks and add masks that utilize these landmarks, which are necessary for our LOM as illustrated in Figure~\ref{occ_train_dataset}.

This section details the operation of the Landmark Occlusion Module. Figure~\ref{occ_results} demonstrates the module's ability to accurately forecast whether a face is partially occluded and the precise location of the occlusion. Notably, the results obtained from the occlusion prediction module will be employed for segmentation based on the distance between different regions and the occluded points. (Please refer to Section \ref{SEGM} for further information.)

The modified network for facial landmark location and landmark occlusion, based on the PFLD work \cite{guo2019pfld}, is depicted in Figure~\ref{landmark_occ_module}. We first use the trained landmark location model and apply it to improve occlusion predictions. An accurate, efficient, and compact facial landmark detector and occlusion landmark detector are essential for segmentation training. We also use the PPM (Section~\ref{PPM}) to aid the training process. Finally, the function $F(\text{landmark}, \text{Occlusion}, \text{Mask})$ judges the facial part occlusion, giving more weight to the occluded regions in the segmentation training process.


\subsection{De-Occlusion Module} \label{DOM}
We obtain accurate regions of occluded parts from the LOM. If the corresponding label in the training data is missing, we use the de-occlusion module to obtain the occlusion mask and incorporate it into the \textsl{SEGM} along with face part mask labels. We tackle the task of face de-occlusion for various types of occlusions using open test and train sets such as FaceOccMask-HQ  and CelebAMaskOcc-HQ Dataset.

In this work, our objective is to address the problem of recovering an occlusion-free face from its noisy observation with occlusion, using the architecture shown in Figure~\ref{de-occlusion_moudle}. Specifically, we propose a multi-scale spatial LSTM encoder that learns representations from the input occluded face $X^{\text{occ}}$, where $X^{\text{occ}}$ denotes the occluded face, and the occlusion-free face $X$. Our goal is to find a function $f$ that removes the occlusion from $X^{\text{occ}}$ by minimizing the difference between the recovered face $f(X^{\text{occ}})$ and the occlusion-free face $X$:
\begin{align}
\label{eq:kmeansObj}
\sum_{i=1}^n \min_{j\in\{1,\dots ,k\}} ||f(X^{\text{occ}}) - X||^2_F
\end{align}
During the encoder processes, we divide the image into $MXN$ patches to mitigate the negative impact of occlusion, as depicted in the left panel of Figure~\ref{de-occlusion_moudle}. These patches are sequentially fed to a spatial LSTM network, which is an extension of the LSTM model designed to analyze two-dimensional signals. Once all patches have been fed, the spatial LSTM encoder outputs its last hidden state in the sequence as a feature representation of the occluded face. This representation is then decoded recurrently to extract face and occlusion information for face recovery.

In the LSTM decoder, we aim to map the learned representation back into an occlusion-free face. As explained earlier, while some types of noise are spatially contiguous, various occlusions on the face in the real scene corrupt the image more maliciously. Therefore, we design a step-by-step contiguous approach to recover the occlusion-free face, as shown in Figure~\ref{de-occlusion_moudle}. The De-Occlusion module ensures the progressive and effective recovery of occluded facial parts to improve face segmentation.

\begin{figure*}[htp]
\centering
\resizebox{0.8\textwidth}{!}{\includegraphics{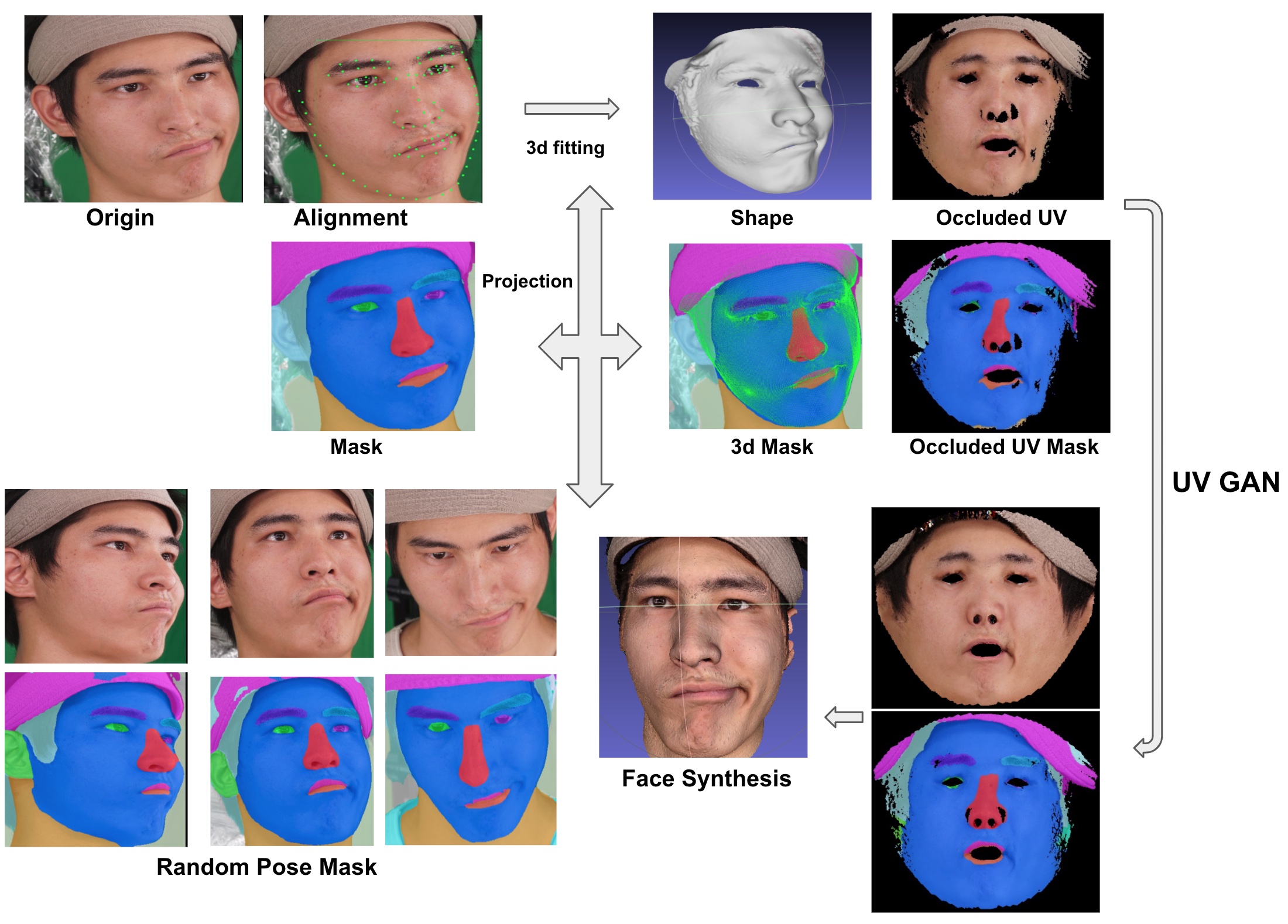}}
\caption{The employed UV GAN module. When fitting a 3DMM to the image, we take the occluded UV map and occluded UV mask map, and we use a generative model to get the complete UV map corresponding UV part mask. By randomly rotating the 3D shape, we get the 2D face with an arbitrary pose with the parsing mask. In this way, we augment pose estimation in the training process.}
\label{UVM}
\end{figure*}

\subsection{UV GAN Module} \label{SectionUVM}
This module utilizes adversarial facial UV map completion~\cite{deng2018uv}. During the occlusion handling process, the UV map completion takes advantage of the face's symmetry and produces a highly realistic UV map. Using the 3D face shape, we can obtain more faces from different angles and correspond them to 2D faces. In the module, as shown in Figure~\ref{UVM}, we first collect the same face from different angles and obtain the self-occluded UV map. Then, we combine the self-occluded UV map and the whole UV map to train a generative model that can inpaint the occluded UV map. We use a robust inpainting UV GAN model to generate our face parsing datasets. The basic training methods can be found in \cite{deng2018uv}. In this way, we obtain correspondent face part masks with pose rotation. Finally, the UV GAN module provides a semi-supervised augmentation for the face parsing training process.


\subsection{Pose Prediction Module (PPM)} \label{PPM}
Head pose estimation is a technique based on deep learning that has been widely used in various applications such as pattern recognition and intelligent human-computer interaction. It is a fundamental task in computer vision and is often used in applications such as facial landmark detection, gaze detection, and other related head pose estimation tasks.

Focusing on object recognition in images, \cite{ruiz2018fine} proposed a method to predict head rotation from image intensities in a direct, accurate, and robust manner. The proposed method is helpful for segmentation and landmark location tasks. The pose estimation is directly trained based on a learning model, which uses Euler angles to describe the three angles of the rigid head direction. However, large pose changes often result in greater errors, so we use the pose estimation to assign more weight to large poses in the segmentation training process.

\subsection{Segmentation Module} \label{SEGM}
We selected a light network, called \textit{RTFormer}~\cite{wang2022rtformer}, as it allows for combining both local and global features of the face in the segmentation process. We propose a segmentation module based on RTFormer, which is used as a segmentation baseline. We then use the results of the previous modules to enhance the robustness of the overall segmentation, particularly in the occluded regions and areas close to the occlusion, where semantic segmentation is often challenging. The occlusion points are obtained from LOM (Section\ref{LOM}), while PPM (Section~\ref{PPM}) estimates the head pose when the head is rotated at a large angle. To support this, we design a loss function for the segmentation:
    \begin{align}
        \begin{aligned}
        & \underset{\mathbf{w}, b, e}{\text{minimize}}
        & & \frac{1}{N}\|\mathbf{w_{CE})}\|_2^2 + C\left(\sum_{i = 1}^N ((x_i-x)o_ip_i)^2\right)\\
        & \text{subject to}
        & & y^{(i)}(\mathbf{w}^T\mathbf{x}^{(i)} + b) \geq 1 - e_i, \, \, \forall \; i = 1, \dots, N\\
        & & & e_i \geq 0, \, \, \forall \; i = 1, \dots N
        \end{aligned}
    \end{align}
    

To ensure optimal segmentation performance, we minimize the following loss function. The term $w_{ce}$ represents the averaged cross-entropy among all the segmentation networks. We also incorporate the distance $o_i$ between the face point and the occluded part, as well as the head pose ratio $p_i$. This supervised hyper-parameter guides the segmentation and its performance is illustrated in Figure~\ref{face_parsing_results}.

\begin{figure}[ht]
\centering
\resizebox{1.0\columnwidth}{!}{\includegraphics{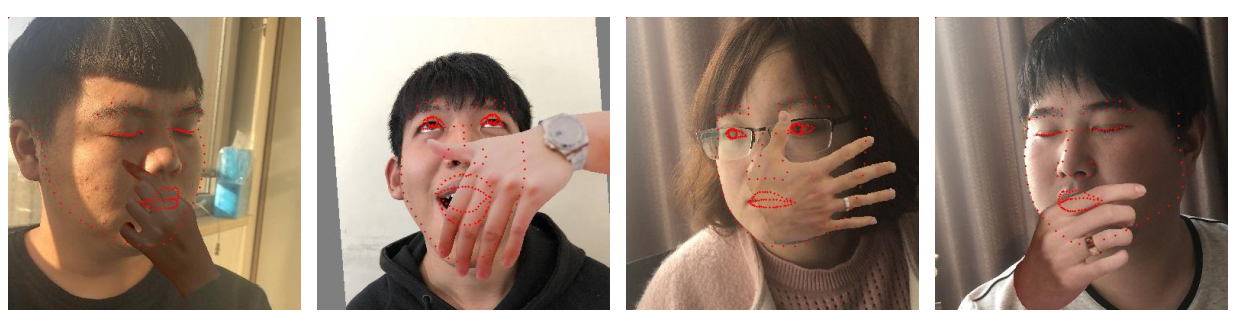}}
\caption{FaceOccMask-HQ, they have the dense points and common occlusion part with its mask.}
\label{264point_occ}
\end{figure}

\begin{figure}[ht]
\centering
\includegraphics[scale=0.20]{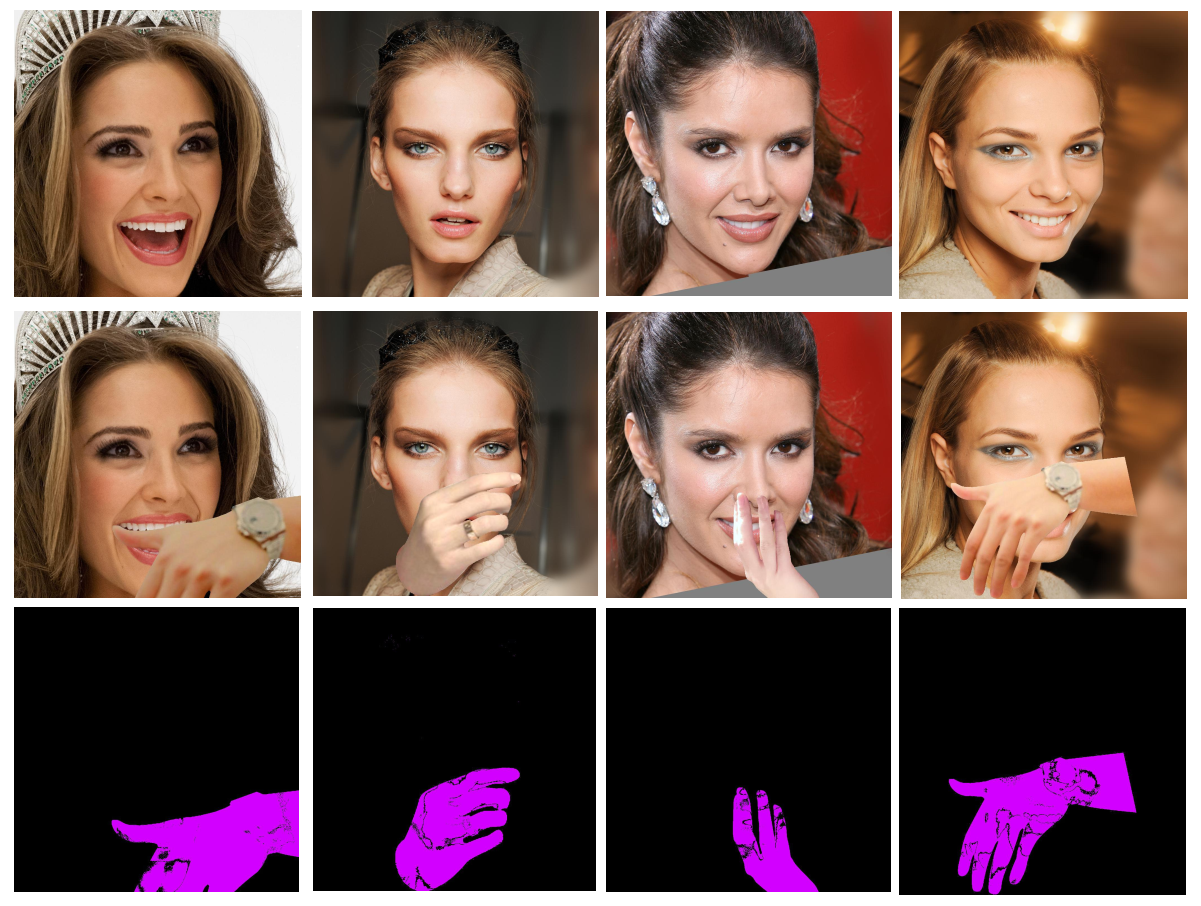}
\caption{CelebAMaskOcc-HQ dataset, they have their own label including some common occlusion we added.}
\label{CeleAMaskOcc}
\end{figure}

To illustrate the relationship between the de-occlusion DOM (Section~\ref{DOM}) and SEGM (Section~\ref{SEGM}), we present the results of the de-occlusion, which produces a mask that uses the occlusion-free face and the original face, as shown in Figure~\ref{de-occlusion_moudle}. This provides a complete view of the efficiency of occlusion-free face parsing and occlusion mask generation. When testing the segmentation model's performance, we can use a single segmentation model to test our dataset, which still yields good results. Additionally, by combining the UV GAN module (see Figure~\ref{UVM}) with the segmentation model, we can achieve even better performance in face parsing.


\section{Experiments}
\subsection{Dataset} \label{Dataset}

We use two datasets in our experiments, namely FaceOccMask-HQ in Figure~\ref{264point_occ} and CelebAMaskOcc-HQ in Figure~\ref{CeleAMaskOcc}. To build our face parsing dataset, FaceOccMask-HQ, we label 264 facial points according to different facial parts such as eyes, mouth, eyebrows, iris, and others, as shown in Figure~\ref{264point}. During the training process, we apply data augmentation techniques such as random background replacement with non-face images or pure colors, random rotation and scaling around the face center, random horizontal flipping, and random gamma adjustment. Our proposed network achieves an efficiency of 44ms per face on an Nvidia 1080ti GPU, which we evaluate in terms of performance.

Further, we utilized the CelebAMask-HQ dataset to create the CelebAMaskOcc-HQ dataset (shown in Figure~\ref{CeleAMaskOcc}). During this process, we used dense points to define specific areas for occlusion augmentation. Each image in this dataset is annotated with 19 labels, including skin, nose, eyes, eyebrows, ears, mouth, lip, hair, hat, eyeglass, earring, necklace, neck, cloth, and occlusion. Additionally, the inserted occlusions in the dataset include hands, masks, and microphones, which makes it useful for further research on these topics.

\begin{table}
\caption{MIOU of Mask FPAN variants and baseline.}
\label{CeleAMaskOcc-HQ all module}
\begin{center}
\begin{tabular}{@{}l|c c c c@{}}
\hline
Method &eyes &mouth &occlusion  &overall\\
\hline\hline
SEGM         & 0.7041 & 0.7212 & 0.3431  &0.7353\\
SEGM+PPM     & 0.7121 & 0.7345 & 0.4021  &0.7521\\
SEGM+PPM &  &  &   &\\
\phantom{SEGM}+LOM & 0.7445 & 0.8205 & 0.6243  &0.8052\\
SEGM+PPM &  &  &   &\\
\phantom{SEGM}+LOM &  &  &   &\\
\phantom{SEGM}+DOM & 0.8015 & 0.8305 & 0.8523  &0.8453\\
Mask FPAN    & 0.8402 & 0.8815 & 0.9203 &\textbf{0.9013}\\
\hline
\end{tabular}
\end{center}
\end{table}

Additionally, we use the FaceOccMask-HQ dataset for evaluating common segmentation methods, while the CelebAMaskOcc-HQ dataset is used for evaluating segmentation under occlusion. Furthermore, it is important to note that the performance of all evaluation metrics on these datasets is heavily dependent on the performance of other modules, primarily landmark occlusion localization and de-occlusion estimation.

\subsection{Ablation Study}

\begin{table}
\caption{Comparison with state-of-the-art methods on FaceOccMask-HQ.}
\label{FaceOccMask-HQ_comparsion}
\begin{center}
\begin{tabular}{@{}l|c c c c@{}}
\hline
Method &eyes &mouth  &occlusion & overall\\
\hline\hline
Luo\cite{luo2012hierarchical} & 0.764 & 0.702  & - &0.802\\
Liu\cite{liu2017face} & 0.782 & 0.748  & - & 0.831\\
Ours & 0.846 & 0.885 &0.932 &0.901\\
\hline
\end{tabular}
\end{center}
\end{table}

\begin{table}
\caption{Comparison with state-of-the-art methods on CeleAMaskOcc-HQ.}
\label{CeleAMaskOcc-HQ_comparsion}
\begin{center}
\begin{tabular}{@{}l|c c c c@{}}
\hline
Method &eyes &mouth  &occlusion & overall\\
\hline\hline
Luo\cite{luo2012hierarchical} & 0.725 & 0.687 & - &0.782\\
Liu\cite{liu2017face} & 0.762 & 0.737  & - & 0.801 \\
Ours & 0.821 & 0.832  & 90.1 &0.848\\
\hline
\end{tabular}
\end{center}
\end{table}

\begin{figure*}[htp]
\centering
\resizebox{0.9\textwidth}{!}{\includegraphics{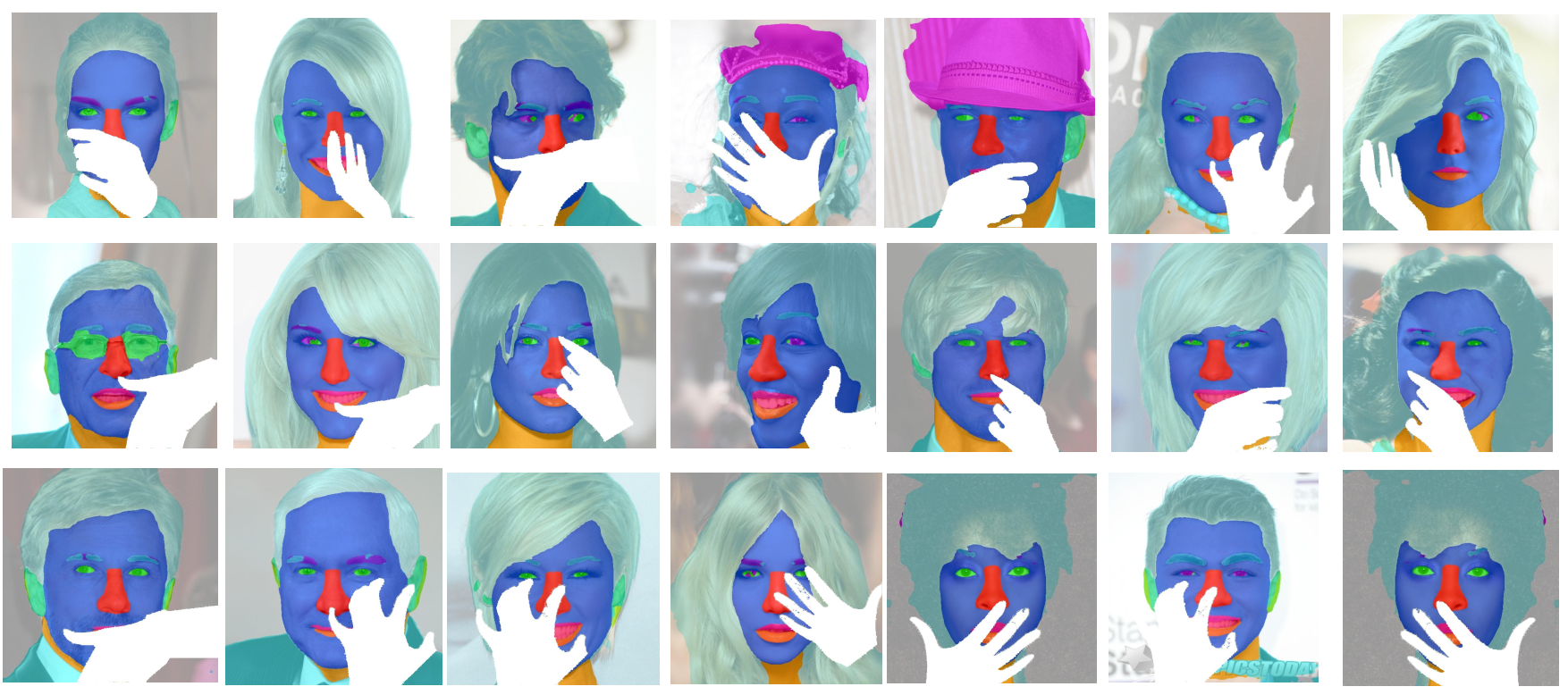}}
\caption{Our method can handle face segmentation with hand occlusions.}
\label{face_parsing_results}
\end{figure*}

As shown in Table~\ref{CeleAMaskOcc-HQ all module}, our MaskFPN framework with fused modules achieves significantly better results than the framework without any other modules (with MIOU increasing from 0.7353 to 0.9013 on the CelebAMaskOcc-HQ dataset). Moreover, as shown in Figure~\ref{framwork}, SEGM + PPM, SEGM + PPM + LOM and SEGM + PPM + LOM + DOM(MASK FPAN) all improve the performance, demonstration that each facial information can makes its contribution to get precise semantic segmentation. Since we have augmented the datasets to include more occlusion parts, it should be noted that the face part of the dataset has more than 19 parts, as mentioned before, including various types of occlusion parts. For the convenience of readers, we have recorded several representative parts in Table~\ref{CeleAMaskOcc-HQ all module}.

\begin{figure}[ht]
\centering
\resizebox{0.93\columnwidth}{!}{\includegraphics{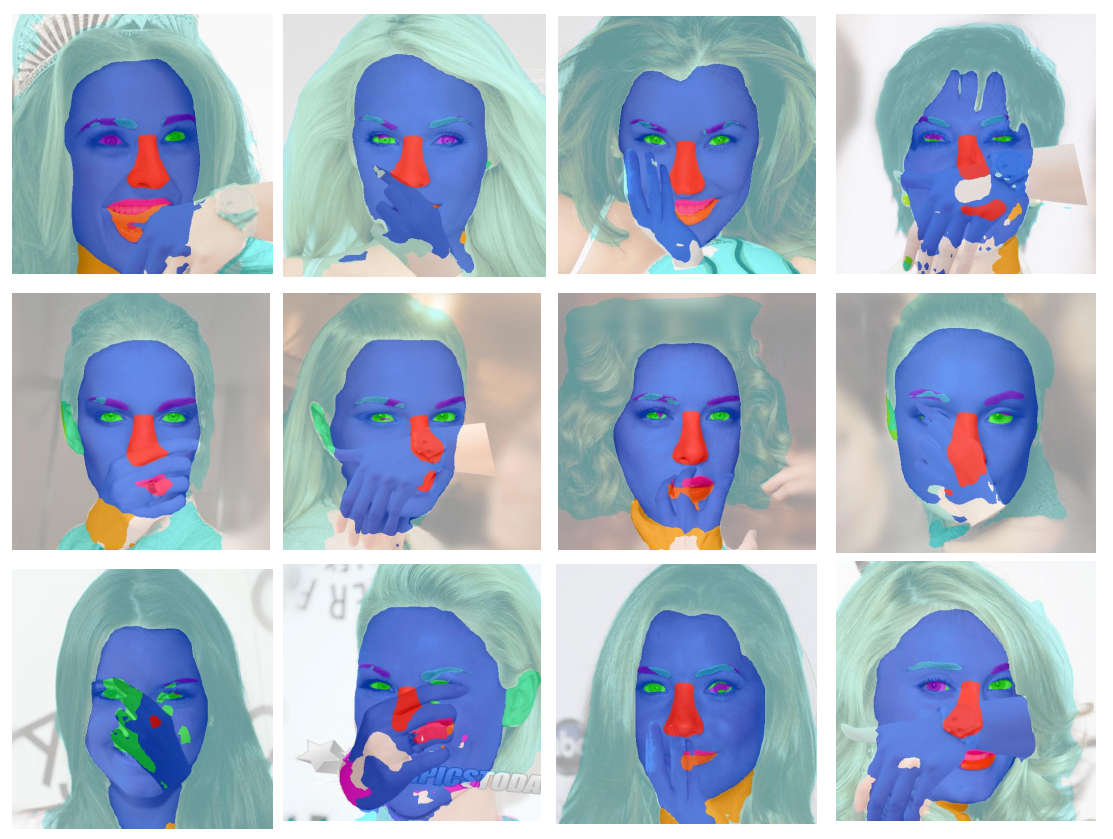}}
\caption{Shown are the original parsing results when faces are occluded, which inspired us to use the de-occlusion to obtain semi-supervision face parsing.}
\label{parsing_without_de_occ}
\end{figure}

\begin{figure}[htp]
\centering
\resizebox{0.9\columnwidth}{!}
{\includegraphics{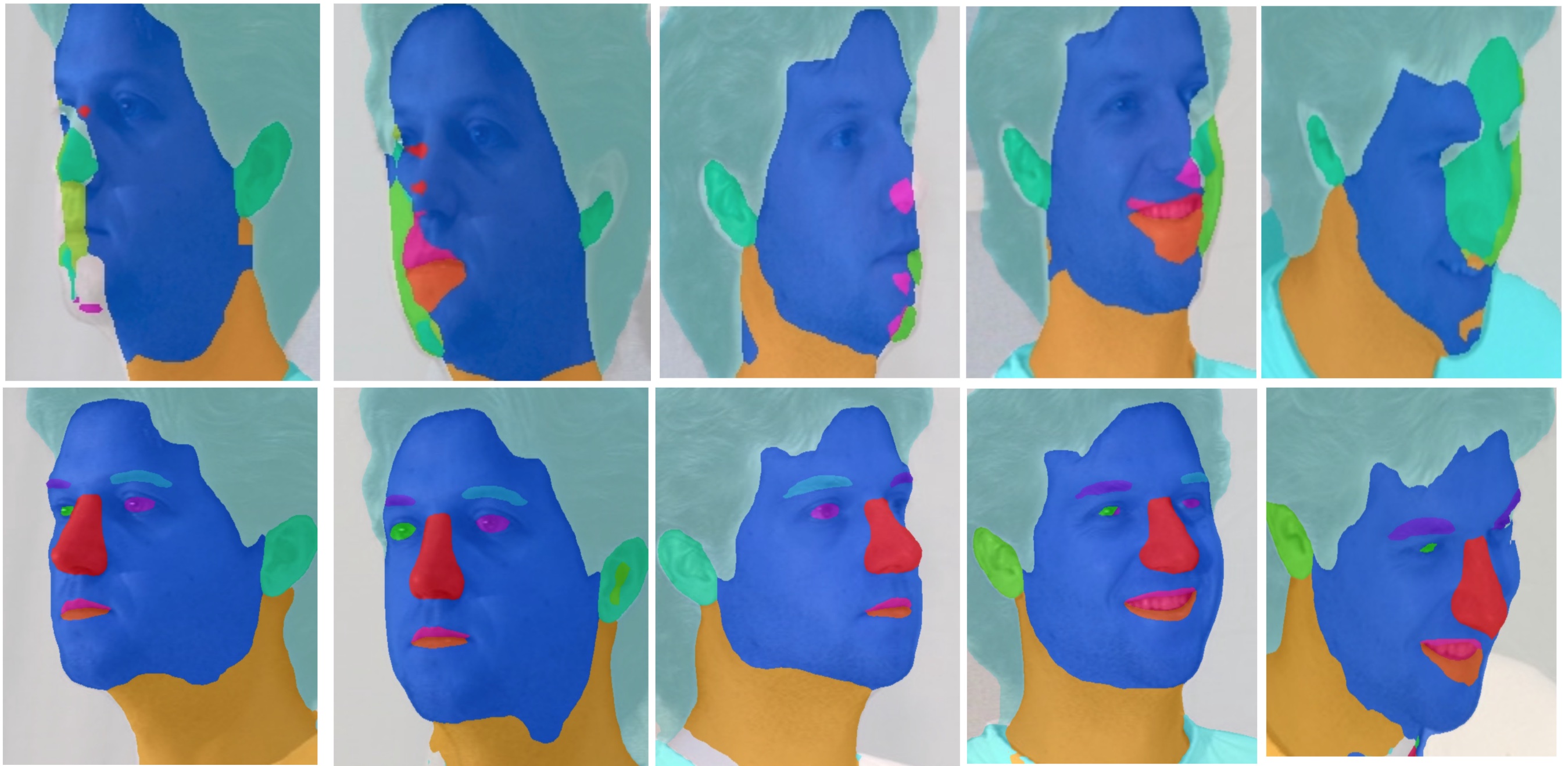}}
\caption{Our method can handle face parsing with large pose variations.  Our method can handle arbitrarily large angles because of the UV GAN module. The first row is the direct segmentation task and the second row is the result of adding de-occlusion and UV gan, etc.}
\label{pose_parsing}
\end{figure}

\subsection{Evaluation} \label{Evaluation}

Our goal is to obtain accurate face parsing even in the presence of occlusions and large pose variations. Figure~\ref{face_parsing_results} demonstrates that our proposed method provides a good solution for limited occlusion parts, and also generates good results in regions close to the occluded regions. The figure shows that the face with occlusions is able to identify the location of the occlusion and generate an occlusion mask. Furthermore, Figure~\ref{pose_parsing} compares the qualitative results between direct segmentation results and our proposed method. The first row shows poor segmentation of facial parts under large pose variations, which inspired us to develop a method to alleviate the influence of large poses, as shown in the second row.

We conducted a comprehensive comparison between our proposed model and the existing state-of-the-art methods on the FaceOccMask-HQ and CelebAMaskOcc-HQ datasets. The evaluation metrics used for measuring the face parsing performance were F-measure, which is commonly used in this domain. Our results were obtained from the former modules of our model. We present the comparison results on the FaceOccMask-HQ dataset in Table~\ref{FaceOccMask-HQ_comparsion}, focusing on the regions that are difficult to distinguish, such as eyes, mouth, hair, and occlusion targets. Furthermore, we provide a comparison of our model with other face parsing methods on the CelebAMaskOcc-HQ dataset in Table~\ref{CeleAMaskOcc-HQ_comparsion}.

Otherwise, the most important aspect of our work is the de-occlusion module, which addresses the problem of occluded facial regions when using a single segmentation method. Figure~\ref{parsing_without_de_occ} demonstrates the poor performance of using a standard segmentation module on the occlusion dataset. While some methods can perform good semantic segmentation on the non-occluded parts of the face, they do not consider the occluded regions, leading to poor segmentation results in the adjacent regions. In contrast, our framework (as shown in Figure~\ref{face_parsing_results}) shows a significant improvement by accurately identifying the occlusion region and providing a precise mask of the occlusion.

As shown in Table ~\ref{FaceOccMask-HQ_comparsion} and Table~\ref{CeleAMaskOcc-HQ_comparsion}, our method surpasses state of the art about face parsing literature. Through qualitative and quantitative analysis, our proposed framework shows state-of-the-art performance for face parsing in the wild. Additionally, because of our light segmentation network, face parsing estimation can be embedded on mobile devices for common use.

\section{Conclusion}
The proposed work presents a novel approach to predict face parsing in the wild, addressing challenges related to occlusion and large pose variations. The approach combines facial landmarks, face occlusion estimations, face de-occlusions, and facial UV GAN creations. The use of occlusion region as attention is inspired by human vision, where attention is drawn towards occluded parts to infer related head and face attributes. A de-occlusion module that employs semi-supervision is proposed to address the problem of occlusion. To leverage the symmetry of the face and the characteristics of its 3D representation, a UV GAN with a 3D morphable model is used to handle face parsing with multiview images, extending to 3D face part segmentation. Additionally, new datasets for face parsing analysis are proposed, and the whole process is utilized to enable semi-supervised face parsing, significantly improving accuracy for custom requirements.



\end{document}